\documentclass[letterpaper, 10 pt, conference]{ieeeconf}  

\IEEEoverridecommandlockouts         
\usepackage{graphicx}
\usepackage{bm}  
\usepackage{ntheorem}   
\usepackage{amsmath} 
\usepackage{amssymb} 
\usepackage{cancel}

\newtheorem{thm}{Theorem}

\newtheorem*{cor}{Corollary}
\newtheorem{defn}{Definition}
\newtheorem*{lem}{Lemma}
\newtheorem*{prop}{Proposition}

\newcommand{\dv}[2]{{\frac{\partial #1}{\partial #2}}}

\newcommand{\dvv}[2]{{\frac{\partial^2 #1}{\partial {#2}^2}}}

\title{\LARGE \bf
On the Covariance of ICP-based Scan-matching Techniques}

\author{Silv{\`e}re Bonnabel, Martin Barczyk and Fran{\c c}ois Goulette
\thanks{Silv{\`e}re Bonnabel and Fran{\c c}ois Goulette are with MINES ParisTech, PSL - Research University, Centre de robotique, 60 Bd St 
Michel 75006
Paris, France
        {\tt\small \{firstname.lastname\}@mines-paristech.fr}}%
        \thanks{Martin Barczyk is with the Department of Mechanical Engineering, University of Alberta, Edmonton AB, T6G 1H9, Canada
        {\tt\small martin.barczyk@ualberta.ca}}%
}

\begin{document}

\maketitle
\thispagestyle{empty}
\pagestyle{empty}

\begin{abstract}
This paper considers the problem of estimating the covariance of roto-translations computed by the Iterative Closest Point 
(ICP) algorithm. 
The problem is relevant for localization of mobile robots and vehicles equipped with depth-sensing cameras (e.g.,~Kinect) or Lidar 
(e.g.,~Velodyne). The closed-form formulas for covariance proposed in previous literature generally build upon the fact that the solution to ICP 
is obtained by minimizing a linear least-squares problem. In this paper, we show this approach needs caution because the rematching step of 
the 
algorithm is not explicitly accounted for, 
and applying it to the point-to-point version of ICP leads to completely erroneous covariances. We then provide a formal mathematical proof why the 
approach is valid in the point-to-plane version of ICP, which validates the intuition and experimental results of practitioners.
\end{abstract}

\section{Introduction}
\label{sec:introduction}

This paper considers the covariance of relative roto-translations obtained by applying the well-known Iterative
Closest-Point (ICP) algorithm~\cite{BM92,CM92} to pairs of successive point clouds captured by a scanning sensor moving through a
structured environment. This so-called scan matching~\cite{LM94,SHT09,JG15} is used 
in mobile robotics, and more generally autonomous navigation, to 
incrementally compute the global pose of the vehicle. The resulting estimates are 
typically fused with other measurements, such 
as odometry, visual landmark detection, and/or GPS. In order to apply probabilistic filtering and sensor fusion techniques such as the extended Kalman filter 
(EKF) e.g.~\cite{NBN07,MRRMP10}, EKF variants~\cite{HBG12,BBDG15}, particle filtering methods, or optimization-based smoothing techniques to find a maximum 
likelihood 
estimate as in Graph SLAM~\cite{grisetti2010tutorial}, the probability distribution of the error associated to each sensor is required. 
Since these errors are typically 
assumed to be zero-mean and normally distributed, only a covariance matrix is needed. 

Contrarily to conventional localization sensors, the covariance of relative roto-translation estimates will not only depend on sensor noise characteristics, 
but also on 
the geometry of the environment. Indeed, when using ICP for scan matching, several sources of errors come into play:
\begin{enumerate}
\item the presence of geometry in one scan not observed in the subsequent one(s), that is, lack of overlapping.
\item mismatching of points, that is, if scans start far from each other the ICP may fall into a local (not global) minimum, yielding an 
erroneous roto-translation 
estimate.
\item even if 1) and 2) do not occur, the 
computed estimate will still possess uncertainty due to sensor noise and possibly underconstrained environments, such as a long featureless corridor.
\end{enumerate}
In practice, the first problem can be addressed by rejecting point pairs with excessive distance metrics or located close to the scanning 
boundaries~\cite{RL01}, and the second by using dead reckoning estimates to pre-align scans or employing a sufficiently fast sampling rate. We will thus focus 
on 
the third source of error.

The covariance of estimates obtained from a scan matching algorithm (such as ICP) can be obtained as follows 
\cite{chowdhury2003stochastic,Cen07}. The estimated transformation $\hat x$ 
output by 
the algorithm is defined as a local argmin of a cost function $J(x,z)$ of the transformation $x$ and the data $z$ (scanned point clouds). 
As a 
result we 
always 
have $\dv{}{x}J(\hat x,z)= 0$. By the implicit function theorem $\hat x$ is a function of the data $z$ around this minimum. Because 
of the above identity, a small variation $\delta z$ in the data will imply a small variation $\delta x$ in the estimate as 
$\dvv{J}{x}\delta x+\frac{\partial^2 J}{\partial z\partial x}\delta z=0$, so that $\delta x=-(\dvv{J}{x})^{-1}\frac{\partial^2 J}{\partial z\partial x}\delta 
z$. As a result if $\delta z$ denotes the (random) discrepancy in the measurement due to sensor noise, the 
corresponding 
variability in the estimates $\delta x=\hat{x}-x$ gives $\mathbb{E}(\delta x \delta x^T)=\text{cov}(\hat x)$ as
\begin{equation}
\text{cov}(\hat x):=\left(\dvv{J}{x}\right)^{-1}\frac{\partial^2 J}{\partial z\partial x}\text{cov}(z)\frac{\partial^2 
J}{\partial z\partial x}^T\left(\dvv{J}{x}\right)^{-1}
\label{fghdd}\end{equation}
Our goal is to point out the potential lack of 
validity of this formula for ICP covariance computation, but also to characterize situations where it can safely be used. Specifically, the problem 
with~\eqref{fghdd}
is that it relies on $\delta x=-(\dvv{J}{x})^{-1}\frac{\partial^2 J}{\partial z\partial x}\delta z$, which is based on the local implicit 
function theorem, and which only holds \emph{for 
infinitesimal variations} $\delta x,\delta z$. In the case of ICP, infinitesimal means sub-pixel displacements. Indeed when matching scans, the rematching 
step performed by the ICP makes the cost function far from smooth, so that  the  Taylor expansion 
\begin{equation}\dv{J}{x}(\hat x+\Delta x,z)=\dv{J(\hat x,z)}{x}+ \dvv{J(\hat x,z)}{ x} \Delta x+ O(\Delta x^2)
\label{fgh}
\end{equation}
which is true in the limit $\Delta x\to 0$ may turn out to be \emph{completely wrong} for displacements $\Delta x$ larger than only a few pixels. An example of 
this will be given in Section~\ref{sec:simpleexample}. On the other hand, if the registration errors are projected onto a reference surface as in 
point-to-plane ICP~\cite{CM92}, Equation 
\eqref{fghdd} will provide valid results. This will be formally proven in Section~\ref{sec:Mathproof}.

Our paper is an extension and rigorous justification of the results of \cite{Cen07} and \cite{GIRL03}. Our main contributions are 
to point out the potential shortcomings of a blind application of~\eqref{fghdd} to point-to-point ICP in Section~\ref{sec:mathframework}, and then to provide 
a formal
mathematical proof based on geometry arguments in Section~\ref{sec:Mathproof} for the validity of~\eqref{fghdd} for point-to-plane ICP. Finally the results 
are illustrated on a simple 3D example in 
Section~\ref{gc}.

\section{Mathematical framework}
\label{sec:mathframework}

Consider using ICP for scan matching (either in 2D or 3D). We seek to find the transformation between two clouds of points $\{p_k\}_{1\leq k\leq N}$ 
and 
$\{q_i\}_{1\leq i\leq M}$. This transformation is $X=(R,p)$, a roto-translation such that the action of $X$ on a vector $p$ is $Xp := Rp + T$. Let 
$\pi(k,X)$ 
denote the label $j$ of the point $q_j$ in the second cloud which is the closest to $Xp_k$. The basic 
point-to-point ICP~\cite{BM92} 
consists of the following steps:
\begin{enumerate}
\item initialize $X_{old}=(I,0)$
\item choose a set of $N$ indices $k$ and define $\pi(k,X_{old})$ such that $q_{\pi(k,X_{old})}$ is the point in the second cloud 
which is the closest to $X_{old}\, p_k$
\item find $X_{new}$ as the argmin of $\sum_k||X_{new} \,p_k-q_{\pi(k,X_{old})}||^2$
\item if $X_{new}$ converged to $X_{old}$ then quit, else $X_{old}=X_{new}$ and goto 2) 
\end{enumerate}
We see that the goal pursued by the ICP is to minimize the function
\begin{equation}
 J(\{p_i\},\{q_i\};\sigma,X):=\sum_k||Xp_k-q_{\sigma(k)})||^2\label{ptotp}
\end{equation}
over the roto-translation $X$ and the matching $\sigma: \{1,\cdots,N\}\mapsto  \{1,\cdots,M\}$. As a result, ICP acts as a \emph{coordinate 
descent} which alternatively updates $X$ as \emph{the} argmin $X$ of $ J$ for a fixed $\pi$, and \emph{the} argmin $\pi$ of $ J$ for a fixed $X$ (the latter 
being 
true only for point-to-point ICP). 
This is what allows to prove local convergence of point-to-point ICP as done in~\cite{BM92}. Because of the huge combinatorial problem 
underlying the 
optimization task of jointly minimizing $J$ over the transformation and the matching, ICP provides a simple and tractable (although computationally 
heavy) approach to estimate $X$. The ICP algorithm possesses many variants, such as the point-to-plane version where the cost function is 
replaced by 
\begin{align}
 J(\{p_i\},\{q_i\};\sigma,X):=\sum_k||(Xp_k-q_{\sigma(k)})\cdot n_{\sigma(k)}||^2\label{ptopl}
\end{align}
that is, the registration error is projected onto the surface unit normal vector of the second cloud at $q_{\sigma(k)}$.  An alternative to this in 2D 
exploited 
in  \cite{Cen07} consists of creating a reference 
surface $S_r$ in $\mathbb R^2$ by connecting adjacent points in the second cloud $\{q_i\}$ with segments, and then employing cost function
\begin{equation}
J(S_r,\{p_k\};X):=\sum_k||Xp_k-\Pi(S_r,Xp_k)||^2
\label{censi:eq}
\end{equation}
where $\Pi(S_r,\cdot)$ is the projection onto 
the surface $S_r$. 
\begin{defn} We define the \emph{ICP cost function} as $ J(\{p_i\},\{q_i\};\pi(\cdot,X),X)$, that is, the  error function $ J(\{p_i\},\{q_i\};\sigma,X)$ with 
closest neighbor matching.
\end{defn}
\begin{defn}
\label{defn:stability}
The stability of the ICP in the sense of \cite{GIRL03} is defined as the variation of the ICP cost function  when $X$ moves a little away from the 
argmin $\hat X$. 
\end{defn}
Definition~\ref{defn:stability}'s terminology comes from the theory of dynamical systems. Indeed, the changes in the ICP cost indicate the ability (and speed) 
of the algorithm to 
return to its minimum when it is initialized close to it. If the argmin $\hat X$ is changed to $\hat 
X+\delta X$ and the cost does not change, then the ICP algorithm output will remain at $\hat X+\delta X$ and will not return to its original value $\hat X$. 

Meanwhile, although closely related, the covariance is rooted in statistical considerations and not in the dynamical behavior of the algorithm:
\begin{defn}The covariance of the ICP algorithm is defined as the statistical dispersion (or variability), due to sensor noise, of the transformation 
$\hat X$ computed by the algorithm over a large number of experiments.
\end{defn}

\subsection{Potential lack of validity of \eqref{fghdd}}\label{pointt}

\subsubsection{Mathematical insight}

Consider the   ICP cost function  $ J(\{p_i\},\{q_i\};\pi(X,\cdot),X)$. To simplify notation we omit the point clouds and 
we let $F(X)$ be the function $X\to J(\pi(X,\cdot),X)$.  At convergence we have by construction of the ICP algorithm at the argmin $\partial_2J(\pi(\hat 
X,\cdot),\hat X)=0$, where the $\partial_2$ denotes 
the derivative with respect to the second argument. As the closest point matching $X\to\pi(X,\cdot)$ is locally constant (except on a set of null measure), 
since an infinitesimal change of each point does not change the nearest neighbors except if the point is exactly equidistant to two distinct points, we have 
$$\frac{d^2F}{dX^2}(\hat X)=\partial_2^2 J(\pi(\hat X,\cdot),\hat X)
$$that is, the matching can be considered as fixed when computing the Hessian of  $X\to J(\pi(X,\cdot),X)$ at the argmin. But the first-order approximation 
\begin{align*}&\partial_2J(\pi(\hat X+\delta X,\cdot),\hat X+\delta X)\\&\quad\approx\partial_2J(\pi(\hat X,\cdot),\hat 
X)+\partial^2_2J(\pi(\hat X,\cdot),\hat X)\delta X\end{align*}
can turn out to very poorly model the stability of the algorithm at the scale $\delta X = \Delta X$ of interest to us, precisely because when moving away from 
the 
current argmin, rematching occurs and thus $\pi(\hat X+\Delta X,\cdot)\neq\pi(\hat X,\cdot)$.

Whereas a closed form estimate of $\partial^2_2J(\pi(\hat X,\cdot),\hat X)$ is easy to calculate, obtaining a Taylor expansion around the minimum 
\emph{which accounts 
for rematching} would require sampling the 
error function all around its minimum, leading to high computational cost~\cite{BB03}.

\subsubsection{Illustration} 
\label{sec:simpleexample}

\begin{figure}[h]
   \includegraphics[width=.48\textwidth]{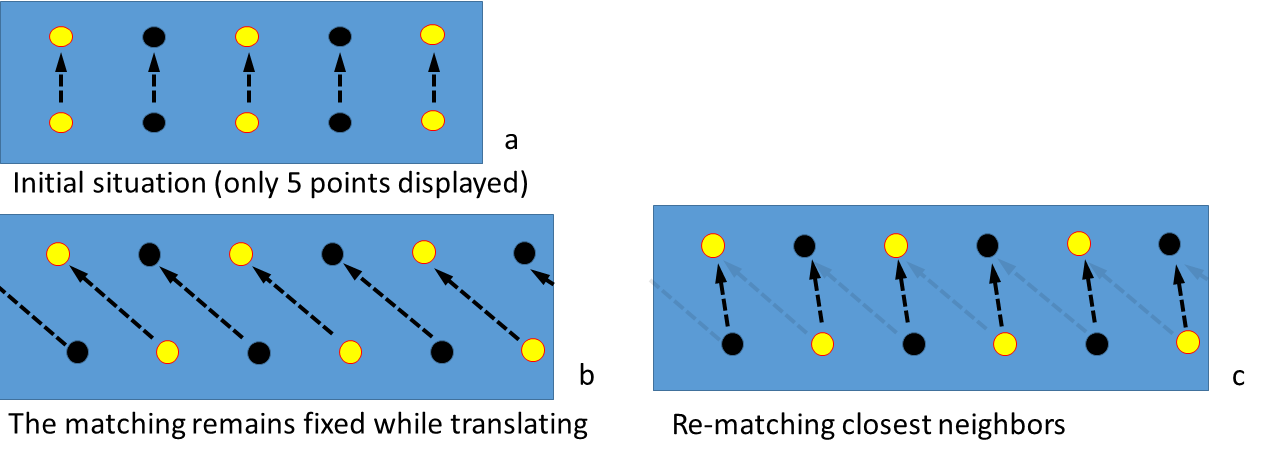}\caption{Point-to-point ICP illustration. (a) The first cloud is made of 10 equally 
spaced 
collinear points (e.g.,~scans of a flat wall), the second cloud is obtained by duplication. To ensure overlap we focus on the central 
points.  (b) While 
translating the second cloud to the right, no re-matching occurs. (c) Translation and re-matching with closest points. The costs $J$ corresponding to cases b
and c are shown in Fig.~\ref{gif}.}\label{gif1}
\end{figure}

\begin{figure}[h]
   \includegraphics[width=.48\textwidth]{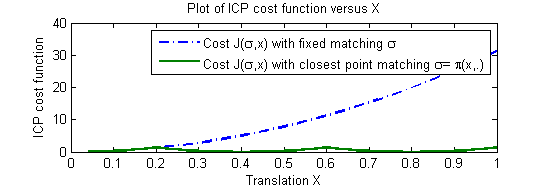}\caption{Point-to-point ICP results from Fig.~\ref{gif1}. Dashed line: plot of the second-order 
approximation to the 
cost $J(\pi(\hat 
X,\cdot),\hat X)+0+\dvv{}{X} J(\pi(\hat X,\cdot),\hat X)(X-\hat X)^2$ versus translations. Due to the quadratic form of the cost 
\eqref{ptotp} the second-order approximation is also equal to $J(\pi(\hat X,\cdot),X)$, i.e.~the cost with matching held fixed (Fig~\ref{gif1} case b).  
Solid line: Plot of the 
true ICP cost $J(\pi(X,\cdot),X)$, i.e. accounting for re-matching with closest points (Fig~\ref{gif1} case c).  }\label{gif}
\end{figure}

Consider a 
simple 2D example of a scanner moving parallel to a flat wall using point-to-point ICP. Figures~\ref{gif1} and~\ref{gif} illustrate the 
fallacy of considering a second-order Taylor expansion of the cost, i.e. computing the cost with  fixed matching. Indeed, Fig.~\ref{gif} displays the 
discrepancy between the true ICP cost and its second-order approximation around the minimum when moving along a 2D wall. We see that rematching 
with closest point correctly reflects the underconstraint/inobservability of the environment, since the cost function is nearly constant as we move along 
the featureless wall. On the other hand, the second-order approximation does not. This proves the Hessian $\dvv{J}{X}$ to the cost at the minimum does not 
correctly reflect the change in the cost function value, and thus the stability of the algorithm. 


Regarding covariance, it is easy to see that Equation \eqref{fghdd} will not reflect the true covariance of the ICP either, as the true covariance should 
be very large (ideally infinite) along the wall's direction, which can only happen if $\dvv{J}{X}$ is very small, but which is not the 
case 
here. 

\subsection{Covariance of linear least-squares}
Consider the linear least-squares minimization problem with cost function
\begin{equation}
J(x)=\sum_i||d_i-B_ix||^2
\label{eq:LScost}
\end{equation}
The solution is of course
\begin{align}
\hat x=\bigg(\sum_i B_i^T B_i\bigg)^{-1}\bigg(\sum_i B_i^T d_i\bigg)
\label{fluc:eq}\end{align}
Let $A:=(\sum_iB_i^TB_i)$, which represents the (half) Hessian $\frac{1}{2} \partial_2^2 J$ of the cost function $J$. Note that $A^T=A$. 
If the measurement $d_i$ satisfies $d_i=B_ix+w_i$ where $x$ is the true parameter and $w_i$ a noise, the covariance of the least squares estimate over a great 
number of experiments is
\begin{align}
\text{cov}(\hat x)&= \mathbb E\left\langle (\hat{x}-x)(\hat{x}-x)^T \right\rangle \nonumber\\
&= \mathbb E\bigg\langle A^{-1}\Big(\sum_i B_i^T w_i\Big)\bigg[A^{-1}\Big(\sum_j B_j^T w_j\Big)\bigg]^T\bigg\rangle \nonumber\\
&=A^{-1}\sum_i\sum_j\Big(B_i^T \mathbb E(w_iw_j^T) B_j\Big)A^{-1}
\label{eq:covfull}
\end{align}
which indeed agrees with \eqref{fghdd}. 
Furthermore, if the $w_i$'s are identically distributed independent 
noises with covariance matrix $\mathbb E(w_iw_j^T)=\sigma^2I\delta_{ij}$, we recover the well-known result \cite[Thm.~4.1]{Kay} that
$$
\text{cov}(\hat x)=\sigma^2A^{-1}\Big(\sum_iB_i^TB_i\Big)A^{-1}=\sigma^2A^{-1}
$$
meaning the (half) Hessian to the cost 
function $A$ encodes the covariance of the estimate.

\subsection{Application to point-to-point ICP}
\label{sec:ppmatrices}

The application of the least-squares covariance formulas to point-to-point ICP can be done as follows~\cite{BB01}. Note in the 3D case the roto-translation $X$ 
is a member of $SE(3)$, the Special Euclidean Lie group with associated Lie 
algebra $se(3) \ni \xi$. Using homogeneous coordinates this writes
\begin{align*} X&=\begin{bmatrix} R & p \\ 0 & 1 \end{bmatrix}, \quad R\in SO(3), p \in \mathbb{R}^3 \\
\xi&=\begin{bmatrix} S(x_R) & x_T \\ 0 & 0 \end{bmatrix}, \quad 
x_R\in \mathbb{R}^3, x_T \in \mathbb{R}^3 \end{align*}
where $S(\cdot)$ is the $3\times 3$ skew-symmetric matrix $S(a)^T=-S(a)$ such that $S(a)b=a \times b$, $a,b\in \mathbb{R}^3$. The map $\text{exp}: 
se(3) \to SE(3)$ is the matrix exponential $e^\xi:=I+\xi + (1/2!)\xi^2 + \cdots$. As explained in Section~\ref{sec:introduction}, we can assume the scans to be 
aligned 
by ICP start out close to each other. This means $X \in SE(3)$ is close to identity and $\xi \in se(3)$ is close to zero, such that 
\begin{equation}
X=e^\xi \approx I 
+ \xi \Longrightarrow Xp \approx p + x_R \times p + x_T.
\label{eq:Xapproximation}
\end{equation}
We can thus consider the ICP estimate $\hat{X}$ as parameterized by $\hat{x}=(\hat{x}_R,\hat{x}_T)\in \mathbb{R}^6$. Specifically we define the linear map $L: 
\mathbb{R}^6 \to se(3)$, $L(\hat{x})=\hat{\xi}$ such that $I + L(\hat{x})\approx \hat{X} \ni SE(3)$ for $\hat{X}$ close to identity. The output model $Y_m$ of 
the ICP is then written as
\[ Y_m=\hat{X}=I+L(\hat{x})=I+L(x)+L(\delta x)=X+L(\delta x) \]
where $\delta x=\hat{x}-x$ and $L(\delta x)$ can be viewed as a zero-mean noise term with associated covariance $\mathbb{E}(\delta x \delta 
x^T)=\text{cov}(\hat{x})$ by~\eqref{fghdd}.  

Using~\eqref{eq:Xapproximation} in the point-to-point ICP, the cost function~\eqref{ptotp} with matching fixed at its convergence value 
$\hat{x}$ writes 
\begin{equation*}
 {J}(\pi(\hat{x},\cdot),x) = \sum_i \| p_i + S(x_R) p_i + x_T - q_{\pi(\hat{x},i)} \|^2
\end{equation*}
which can be rewritten in the sum-of-squares form~\eqref{eq:LScost} with
\begin{equation*}
 d_i = p_i - q_{\pi(\hat{x},i)}, \qquad B_i = [S(p_i) \quad -I]
\end{equation*}
The Hessian 
$A=\frac{1}{2}\partial_2^2J=\sum_i 
B_i^T 
B_i $ is then equal to 
\begin{equation*}
  \sum_i \begin{bmatrix} -S(p_i)^2 & S(p_i) \\ -S(p_i) & I \end{bmatrix}
\end{equation*}

\subsection{Application to point-to-plane ICP}\label{rus:sec}

For point-to-plane ICP, the cost function~\eqref{ptopl} with matching fixed at its convergence value $\hat x$ and approximation~\eqref{eq:Xapproximation} is
given by
\begin{equation*}
J(\pi(\hat x,\cdot),x) = \sum_i \big[ (  x_R \times p_i + x_T + p_i - q_{\pi(\hat x,i)} )\cdot n_{\pi(\hat x,i)} \big]^2
\end{equation*}
Using the scalar triple product circular property $(a \times b)\cdot c = (b \times c)\cdot
a$, this can also be rewritten in form~\eqref{eq:LScost} with
\begin{equation*}
 \begin{aligned}
d_i&=n_{\pi(\hat x,i)}^T (p_i - q_{\pi(\hat x,i)}) \\
B_i&=\begin{bmatrix} -(p_i \times n_{\pi(\hat x,i)})^T & -n_{\pi(\hat x,i)}^T \end{bmatrix}
\end{aligned}
\end{equation*}
and the Hessian $A=\frac{1}{2}\partial_2^2J=\sum_i B_i^T B_i $ is equal to
\begin{equation*}
\sum_i
\begin{bmatrix} (p_i \times n_{\pi(\hat x,i)})(p_i \times n_{\pi(\hat x,i)})^T & (p_i \times n_{\pi(\hat x,i)})n_{\pi(\hat x,i)}^T\\
n_{\pi(\hat x,i)}(p_i \times n_{\pi(\hat x,i)}) & n_{\pi(\hat x,i)} n_{\pi(\hat x,i)}^T \end{bmatrix}
\end{equation*}
The above expression models the Hessian of $J(\{p_i\},\{q_i\};x,\pi(\hat x,\cdot))$ and was given in~\cite{GIRL03}, who argue by intuition that it  
models the stability of the point-to-plane ICP algorithm. We will formally prove this fact --- that the point-to-plane Hessian correctly captures the behavior 
of the true ICP cost function 
$J(\{p_i\},\{q_i\};x,\pi(x,\cdot))$ around $\hat x$ --- in Section~\ref{sec:Mathproof}.

\section{A rigorous mathematical result for point-to-plane ICP}
\label{sec:Mathproof}

The present section is devoted to prove that as far as point-to-plane ICP is concerned, and unlike the point-to-point case, Equation~\eqref{fgh} and 
hence~\eqref{fghdd} is 
indeed valid, even for large $\Delta x$. In fact, a bound on $\Delta x$ depending on the curvature of the scanned surface is given, allowing to 
characterize 
the domain of validity of the formula.  This result is novel and provides a rigorous framework to justify the intuitive arguments in~\cite{GIRL03}.

\begin{thm}Consider a 2D environment made of (an ensemble of disjoint) smooth surface(s) $S_r$  having maximum curvature $\kappa$. Consider a cloud of points 
$\{a_i\}$ obtained by scanning the environment. Consider the cost $J(\pi(x,\cdot),x)$  obtained by matching the cloud $\{a_i\}$  with the displaced cloud 
$\{a_i\}+x_R\times\{a_i\}+x_T$  where $x:=(x_R,x_T)$ are the motion parameters. As $0$ is a global minimum the gradient vanishes at $x=0$. 
The following second-order Taylor expansion\begin{equation}
\begin{aligned}&J(\pi(x+\Delta x,\cdot),x+\Delta x)\\&=J(\pi(0,\cdot),0)+\partial_2^2J(\pi(0,\cdot),0) ||{\Delta x}||^2+O(\kappa||{\Delta x}||^3)
\end{aligned}\end{equation}
is \emph{valid} for $\Delta x$ sufficiently small, but large enough to let rematching occur. 
\label{thm}\end{thm}
Note  that if the environment is made of disjoint planes, we have $\kappa=0$ and both cost functions agree \emph{exactly}. The remainder of this section is 
devoted to the proof of the theorem, and a corollary proving the result remains true in 3D. 

\subsection{Details of result}
 The proof of the previous theorem is based on the following.
\begin{prop}Consider the assumptions of Theorem \ref{thm}. Around the minimum $ x=0$ the cost with fixed matching
\begin{align*}
&J(\pi(0,\cdot),x)\\
&= \sum_i \big[ ( a_i+ x_R \times a_i + x_T  - a_{\pi(0,i)} )\cdot n_{i} \big]^2
\end{align*}
differs from the true ICP cost in the following way
\begin{align*}
&J(\pi(x,\cdot),x)\\
&= \sum_i \big[ (  a_i+x_R \times a_i + x_T   - a_{\pi( x,i)} )\cdot n_{i} +\psi_i\big]^2
\end{align*} where the approximation error $\psi_i$ is \emph{already} second order in the function arguments
as $
 | \psi_i | \leq 8\kappa (\|x_R\times  a_i  + x_T \|)^2$ as long as $\kappa |s_i-s_{\pi( x,i)}|\leq 1$ where $s_i$ and $s_{\pi( x,i)}$ denote the curvilinear 
abscissae of the points $a_i$ and $a_{\pi( x,i)}$ along the surface $S_r$.
\end{prop}To begin with, note that the condition $\kappa |s_i-s_{\pi( x,i)}|\leq 1$ is independent of the chosen units as $\kappa |s_i-s_{\pi( x,i)}|$ is 
dimensionless. To fix ideas about the validity condition, assume the environment is circular with an arbitrary radius. The above condition means that the 
Taylor 
expansion is proved valid as long as the displacement yields a rematching with the nearest neighbor at most 1 rad ($57.3^\circ$) along the circle from 
the initial 
point. We see this indicates a large domain of validity. Note that in case where the environment is a line both functions coincide exactly. 

\begin{figure}[hpbt!]
    \center
     \includegraphics[width=1\linewidth]{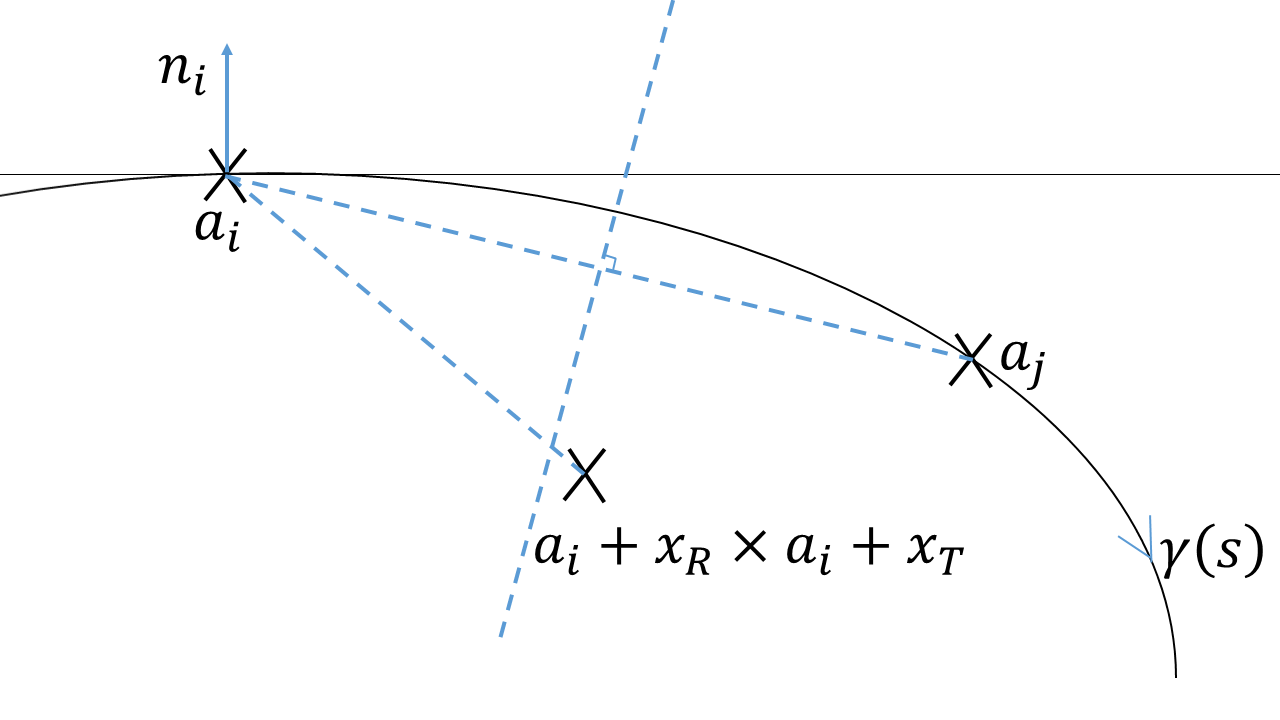}
  \caption{Illustration for the proof.}
\label{fig:proof}
\end{figure}

To prove the result, assume the surface where point $i$ lies is parameterized by $\gamma(s)$ with curvilinear 
abscissa $s$, and with maximum curvature $\kappa$. Such a curve
has tangent vector $\gamma'(s)$ with $\| \gamma'(s)\|=1$ and normal vector $\gamma''(s)$ with $\| \gamma''(s) \|\leq\kappa$, the curvature. The
point cloud $\{a_i\}\in\mathbb{R}^2$ is obtained by scanning this environment at discrete points $\gamma(s_1),\gamma(s_2),\cdots$. 
We assume the surfaces (here curves) are sufficiently disjoint so that under the assumptions of the Proposition, $a_i$ and its closest point $a_{\pi(x,i)}$ 
lie 
on the same curve 
of maximum curvature $\kappa$. 
By writing $ x_R \times a_i + x_T + a_i - a_{\pi( x,i)}= x_R \times a_i + x_T + a_i - a_{\pi( 0,i)}+ a_{\pi( x,i)}$ we see the error
made for each term $i$ is $\psi:=(a_{\pi( 0,i)}- a_{\pi( x,i)})\cdot n_i=(a_i- a_j)\cdot n_i$ where we let $j:={\pi( x,i)}$ and we used the obvious fact that 
$\pi(0,i)=i$. To study $\psi_i$ expand $\gamma(s)$ about $s=s_i$ using Taylor's theorem with remainder:
\begin{equation*}
\gamma(s) = \gamma(s_i) + \gamma'(s_i)(s - s_i ) + \int_{s_i}^s \gamma''({u})(u - s_i )du
\end{equation*}
Take $s=s_{\pi(x, i)}:=s_j$ and project along the normal $n_i$:
\begin{multline*}
(\gamma(s_{j}) - \gamma(s_i))\cdot n_i =(s_{j} - s_i ) \gamma'(s_i)\cdot n_i \\~+  \int_{s_i}^{s_{j}} \gamma''({u})\cdot n_i(u - s_i)du
\end{multline*}
Note $\gamma'(s_i)\cdot n_i=0$ since $\gamma'(s_i)$ is the (unit) tangent vector to the curve at $s=s_i$. Taking absolute values of both sides,
\begin{equation*}
\begin{aligned}  |\psi_i|=| (b_j - b_i) \cdot n_i |&\leq \frac{1}{2} [\max_u\| \gamma''(\tilde{u}) \|] | s_{j} - s_i |^2 \\&\leq \frac{1}{2} \kappa | s_{j} - 
s_i | ^2 \\&\leq \frac{1}{2} \kappa \left(4\|x_R\times a_i+x_T\| \right)^2
 \end{aligned}
\end{equation*}
as claimed. Only the last inequality needs be justified. It stems from the following result:
\begin{lem}
If no rematching occurs, i.e. $i=j$, then $\psi_i=0$. If $i\neq j$, we have for $\kappa|s_i-s_j|\leq 1$ the inequality $|s_i-s_j|\leq 4 \|x_R\times a_i+x_T\|$.
\end{lem}
 Indeed, rematching occurs only if the displaced point $x_R\times a_i+x_T$ is closer to $a_j$, as illustrated in Fig.~\ref{fig:proof}. But this implies the 
distance between the displaced point and $a_i$ is greater than half the distance between $a_i$ and $a_j$ (see Fig.~\ref{fig:proof}), that is $\|a_j-a_i\|\leq 2 
\|a_i+x_R\times a_i+x_T-a_i\|$. Now, another Taylor expansion yields $\gamma(s_j)-\gamma(s_i)=\gamma'(s_i)(s_j-s_i)+\int_{s_i}^{s_j}\gamma''(u)(u-s_i)du$. Using 
$\| \gamma'(s)\|=1$ and $\| \gamma''(s) \|\leq\kappa$ we get 
$\|\gamma(s_j)-\gamma(s_i)\|\geq| s_j-s_i|-\frac{1}{2}\kappa(s_j-s_i)^2\geq\frac{1}{2}| s_j-s_i|$, the latter inequality steming from the assumption that 
$\kappa| s_j-s_i|\leq 1$. Gathering those results we have thus proved$$2 \|x_R\times a_i+x_T\|\geq \|a_j - a_i\|:=\|\gamma(s_j)-\gamma(s_i)\|\geq\frac{1}{2} 
|s_j-s_i|$$which allows to prove the Lemma, and in turn the Proposition. 

\subsection{Extension to the  3D case}
\begin{cor}The results hold in 3D where $\kappa$ denotes the maximum of the Gauss principal curvatures. 
\end{cor} 

The corollary can be proved in exactly the same way as the theorem, by studying the discrepancy between both cost functions term-by-term. The idea is then 
merely to consider the plane spanned by the unit normal $n_i$ and the segment relating $a_i$ and $a_j$. This plane intersects the surface $S_r$ at a 
curve, and the same process can be applied as in the 2D case. The curvature of this curve is by definition less than the maximum Gauss principal curvature of 
the surface. 

\section{Illustration of the results in 3D}
\label{gc}

The covariance of scan matching estimates is computed using Equation~\eqref{eq:covfull}, which requires a model of the measurement noise $w_i$ via its 
covariance $\mathbb{E}(w_i w_j^T)$. Modeling noise of depth sensors is a separate topic and will not be considered in the 
present paper. 
Regardless, it's clear from~\eqref{eq:covfull} that the Hessian $A$ of the cost function plays a key role in this computation. We now demonstrate using a very 
simple numerical example in 3D that the Hessian of the point-to-plane correctly models the behavior of the ICP algorithm.

\begin{figure}[hpbt!]
    \center
     \includegraphics[width=0.9\linewidth]{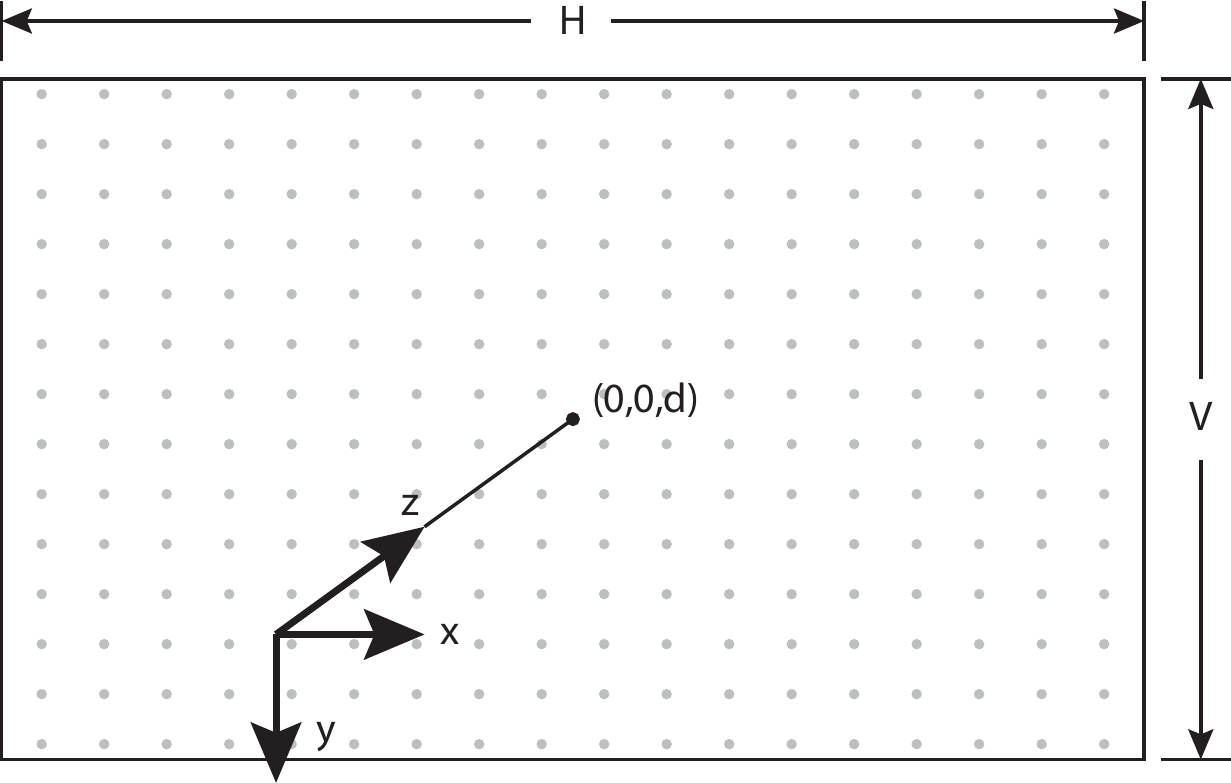}
  \caption{Scan of 3D plane wall with $N_H$ horizontal and $N_V$ vertical points distributed symmetrically about origin.}
\label{fig:3Dwallscan}
\end{figure}

Consider
a 3D scan $\{p_i\}$ of a plane wall by a depth camera located perpendicularly $d$ units away as shown in
Figure~\ref{fig:3Dwallscan}. A depth
image of
$N_H$ by $N_V$ pixels (function of the hardware) captures a surface measuring $H$ by $V$ units (function of the optical field of view and distance $d$) such
that $a_i=[x_i \quad y_i \quad d]^T$ where $-H/2 \leq x_i \leq H/2$, $-V/2 \leq y_i \leq V/2$.

Assume a
previous scan $\{q_i\}$ with associated surface normals $\{n_i\}$ was captured with the same camera orientation at distance $d'$ such that
$q_i=[ x_i' \quad y_i' \quad d']^T$, $n_i= [ 0 \quad 0 \quad -1 ]^T$ where 
$-H'/2 \leq x_i' \leq H'/2$, $-V'/2 \leq y_i' \leq V'/2$. From Section~\ref{rus:sec} we have
\begin{equation*}
\begin{aligned}
 A&=\sum_i \begin{bmatrix}
 y_i^2 & -x_i y_i & 0 & 0 & 0 & y_i \\
-x_i y_i & x_i^2 & 0 & 0 & 0 & -x_i \\
0 & 0 & 0 & 0 & 0 & 0 \\
0 & 0 & 0 & 0 & 0 & 0 \\
0 & 0 & 0 & 0 & 0 & 0 \\
y_i & -x_i & 0 & 0 & 0 & 1 \end{bmatrix} \\
&= \begin{bmatrix}
 \Psi & 0 & 0 & 0 & 0 & 0 \\
0 & \Xi & 0 & 0 & 0 & 0 \\
0 & 0 & 0 & 0 & 0 & 0 \\
0 & 0 & 0 & 0 & 0 & 0 \\
0 & 0 & 0 & 0 & 0 & 0 \\
0 & 0 & 0 & 0 & 0 & N \end{bmatrix}
\end{aligned}
\end{equation*}
where $\sum x_i^2:=\Xi$, $\sum y_i^2:=\Psi$ and $\sum 1:=N$ are non-zero. By
inspection this
$A$ possesses three zero eigenvalues with associated eigenvectors $(e_3,e_4,e_5)\in\mathbb{R}^6$, indicating that in this case rotations
about the $z$
axis and translations along the $x$ and $y$ axes are unobservable to scan matching, which agrees with physical intuition about Figure~\ref{fig:3Dwallscan}. 
Although $A$ is 
singular,~\eqref{eq:covfull} can still be computed by removing $x_3$, $x_4$ and $x_5$ from the state vector $x=[x_R \quad x_T]$, thus
deleting the third, fourth and fifth columns of $B_i$ or equivalently rows and columns of $A$. In this way only the covariance 
of observable 
parameters will be estimated.

Now consider using the point-to-point Hessian given in Section~\ref{sec:ppmatrices}. In this case we have
\begin{equation*}
\begin{aligned}
A&=\sum_i \begin{bmatrix} 
          d^2 + y_i^2 & -x_i y_i & -d x_i & 0 & -d & y_i \\
          -x_i y_i & d^2 + x_i^2 & -d y_i & d & 0 & -x_i \\
          -d x_i & -d y_i & x_i^2 + y_i^2 & -y_i & x_i & 0 \\
          0 & d & -y_i & 1 & 0 & 0 \\ 
          -d & 0 & x_i & 0 & 1 & 0 \\
          y_i & -x_i & 0 & 0 & 0 & 1
         \end{bmatrix} \\
      &= \begin{bmatrix} 
          N d^2 + \Psi & 0 & 0 & 0 & -Nd & 0 \\
          0 & N d^2 + \Xi & 0 & N d & 0 & 0 \\
          0 & 0 & \Xi + \Psi & 0 & 0 & 0 \\
          0 & N d & 0 & N & 0 & 0 \\ 
          -Nd & 0 & 0 & 0 & N & 0 \\
          0 & 0 & 0 & 0 & 0 & N
         \end{bmatrix} \\   
\end{aligned}
\end{equation*}
By inspection this $A$ is full rank and so it does not have zero eigenvalues. Since we know there are three unobservable directions, the 
point-to-point ICP Hessian provides a \emph{completely wrong} model of the scan matching observability (and in turn covariances), exactly as predicted. 

\section{Conclusion}
   
In this paper we have provided a rigorous mathematical proof --- a novel result to the best of our knowledge --- why the closed-form 
formula~\eqref{fghdd} and its linearized version~\eqref{eq:covfull} provide correct roto-translation estimate covariances only in the point-to-plane variant 
of ICP, but not point-to-point.

This paper has not investigated the modeling of the noise term $w_i$ which appears in the linearized covariance formula~\eqref{eq:covfull}. We know that 
assuming this term to be independent and identically distributed Gaussian 
noise will lead to erroneous (overly optimistic) estimates of covariance, as noted in~\cite{BB01} for instance. We are currently investigating how to 
rigorously 
derive a closed-form expression to obtain a valid and realistic covariance matrix for 3D depth sensor-based scan matching. 
  
\section*{Acknowledgments}

The work reported in this paper was
partly supported by the Cap Digital Business Cluster TerraMobilita Project.


\bibliographystyle{IEEEtran}
\bibliography{bibliography}

\end{document}